\documentclass[11pt,a4paper]{article}
\usepackage[hyperref]{acl2021}
\usepackage{times}
\usepackage{latexsym}
\usepackage{multirow}
\usepackage{graphicx}
\usepackage{amsmath}
\usepackage{lipsum}
\usepackage{amsfonts}

\usepackage{microtype}

\newcommand{\squishlist}{
   \begin{list}{$\bullet$}
    { \setlength{\itemsep}{0pt}      \setlength{\parsep}{2pt}
      \setlength{\topsep}{1pt}       \setlength{\partopsep}{0pt}
      \setlength{\leftmargin}{1em} \setlength{\labelwidth}{1em}
      \setlength{\labelsep}{0.5em} } }

\newcommand{\squishend}{
    \end{list}  }

\aclfinalcopy % Uncomment this line for the final submission
\def\modelname{S$^3$E$^2$}

\newcommand\blfootnote[1]{%
\begingroup
\renewcommand\thefootnote{}\footnote{#1}%
\addtocounter{footnote}{-1}%
\endgroup
}

\title{Semantic and Syntactic Enhanced Aspect Sentiment Triplet Extraction}

\author{\text{Zhexue Chen}$^{1,2,3}$ \text{, Hong Huang}$^{1,2,3*}$ \text{, Bang Liu}$^{4*}$ \text{, Xuanhua Shi}$^{1,2,3}$ \text{, Hai Jin}$^{1,2,3}$\\
$^{1}$National Engineering Research Center for Big Data Technology and System\\
$^{2}$Service Computing Technology
and System Lab\\
$^{3}$Huazhong University of Science and Technology, China\\
$^{4}$RALI \& Mila, University of Montreal \\
\texttt{\{chenzhexue,honghuang,xhshi,hjin\}@hust.edu.cn} \\
\texttt{bang.liu@umontreal.ca}}
\date{}

\begin{document}
\maketitle
\blfootnote{*Corresponding Author}
\begin{abstract}
\textit{Aspect Sentiment Triplet Extraction} (ASTE) aims to extract triplets from sentences, where each triplet includes an entity, its associated sentiment, and the opinion span explaining the reason for the sentiment.
Most existing research addresses this problem in a multi-stage pipeline manner, which neglects the mutual information between such three elements and has the problem of error propagation.
%Also, there are triplet extraction models that incorporate a new tagging schema, but they ignore the intrinsic connection between them.
%We discover that there is usually syntactic dependence among triplets and semantically similar words tend to be accompanied by relationships.
In this paper, we propose a \textit {Semantic and Syntactic Enhanced aspect Sentiment triplet Extraction model} (\modelname) to fully exploit the syntactic and semantic relationships between the triplet elements and jointly extract them.
Specifically, we design a Graph-Sequence duel representation and modeling paradigm for the task of ASTE: we represent the semantic and syntactic relationships between word pairs in a sentence by graph and encode it by \textit{Graph Neural Networks} (GNNs), as well as modeling the original sentence by LSTM to preserve the sequential information.
Under this setting, we further apply a more efficient inference strategy for the extraction of triplets.
Extensive evaluations on four benchmark datasets show that \modelname{} significantly outperforms existing approaches, which proves our \modelname's superiority and flexibility in an end-to-end fashion. 
\end{abstract}

\section{Introduction}
\begin{figure*}[t]
\centerline{\includegraphics[width=0.8\textwidth]{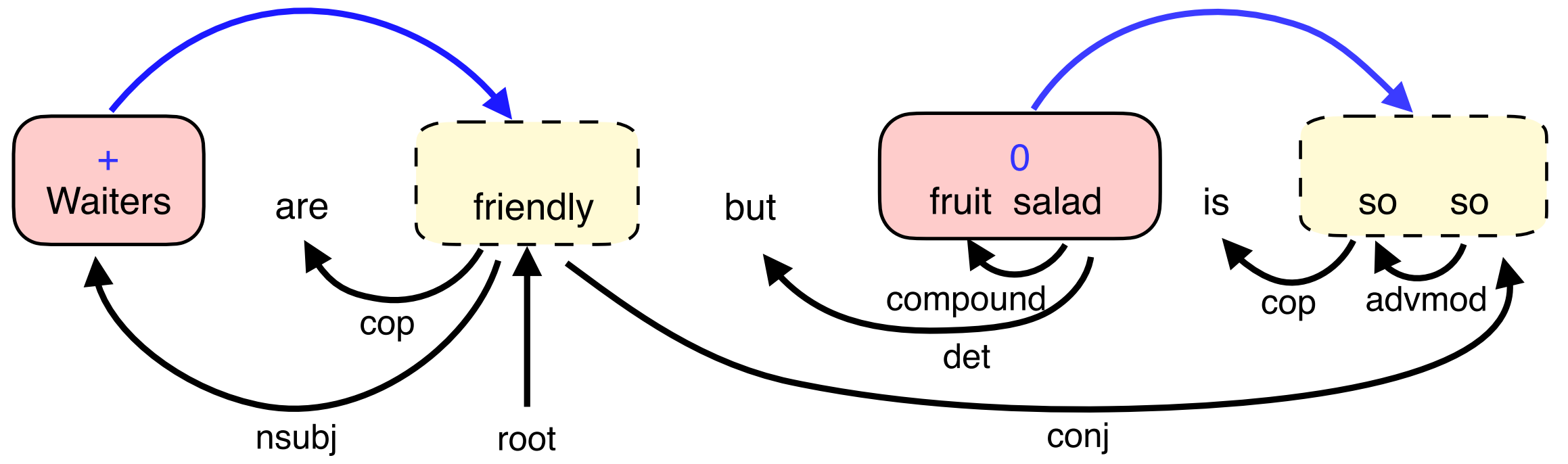}}
\caption{An example of the ASTE task. The words in the solid and dashed boxes are aspects and opinions, respectively. The blue arrows above represent the correspondence between them. The black arrows below represent the dependencies between words.} 
\label{aste_example}
\end{figure*}

%ABSA任务介绍-> ABSA方法的局限性，意义的局限性->提出ASTE->ASTE  example->Peng方法思路和缺点，pipeline->Xu 方法思路和缺点，不能够解决一对多现象(举例句子)->Wu方法思路和缺陷，不能充分利用单词对之间的关系->基于以上观察，我们提出模型, 列举我们的贡献
\textit{Aspect-based Sentiment Analysis} (ABSA) usually requires to extract comment targets in a review and judge corresponding sentiment polarities~\citep{liu2012sentiment,pontiki-etal-2014-semeval}. Such a research field has received widespread attention~\citep{zhang2015neural,AAAI1714931,li2019learning,li2019unified}. In this paper, we concentrate on a more  relatively fine-grained task - \textit{Aspect Sentiment Triplet Extraction} (ASTE)~\citep{Peng_Xu_Bing_Huang_Lu_Si_2020}, which aims to extract triplets, including aspects (e.g., entities), the corresponding sentiment for each aspect, and the opinion spans explaining the reason for the sentiment. An example is shown in Fig.~\ref{aste_example}. It contains two triplets, $\left(Waiters, friendly, +\right)$ and $\left(fruit \ salad, so \ so, 0\right)$ where we use +, -, and 0 to represent positive, negative, and neutral sentiment. Unlike the ABSA task that extracts two tuples, $\left(Waiters, +\right)$ and $\left(fruit \ salad, 0\right)$ in this sentence, such triplets extracted by ASTE task can better reflect multiple emotional factors (aspect, opinion, sentiment) from the user reviews and are more suitable for practical application scenarios.

%\begin{table*}
%\centering
%\resizebox{15.0cm}{1.0cm}{
%\begin{tabular}{c|ccccccccccc}
%\hline
%Method & Input & Waiters & are & friendly & %but & the & fruit & salad & is & so & so\\
%\hline
%\multirow{2}{*} { \citet{Peng_Xu_Bing_Huang_Lu_Si_2020}} & unified tag & S-POS & O & O & O & O & B-NEU & E-NEU & O & O & O \\
%& opinion tag & O & O & S & O & O & O & O & O & B & E\\
%\hline
%\citet{xu2020aspect} & positional tag & B$_{2,2}^{+}$ & O & O & O & O & B$_{3,4}^{0}$ & E & O & O & O \\
%\hline
%Gold triplets & %\multicolumn{10}{c}{(Waiters, POS, %friendly),  (fruit salad, NEU, so so)} \\
%\hline
%\end{tabular}}
%\caption{Comparison of different tagging schema.}
%\label{tab_different_tag_schema}
%\end{table*}

The ASTE task is extremely challenging because it requires extracting these three elements in one shot. Straightforwardly, one naive solution is to split the ASTE task into two stages in a pipeline manner using a unified tagging schema~\footnote{It consists of $\{B,I,E,S\}-\{NEU,NEG,POS\}$ and tag $O$, which denote beginning, inside, end, single-word target with neutral, negative and positive sentiment respectively and outside of a target.}~\citep{Peng_Xu_Bing_Huang_Lu_Si_2020}. Such a pipeline approach lacks an effective mechanism to capture the three elements' relationship and suffers from error propagation. 
Another solution for the ASTE task is to use an end-to-end model to extract triplets~\citep{DBLP:conf/emnlp/XuLLB20,wu-etal-2020-grid}. Yet, these methods focus on designing a new tagging schema to formalize ASTE into a unified task and cannot effectively establish the connection between words and ignore the semantic and syntactic relationship between the three elements. 

Besides, a sentence may contain a one-to-many case, that is, one aspect corresponds to multiple opinions, or one opinion corresponds to multiple aspects. For instance, in the sentence \textit{"We love the food, drinks, and atmosphere,"} the opinion "love" is associated with three aspects ``food'', ``drinks'', and ``atmosphere''. %For another example, the aspect "place" is associated with two opinions in the sentence \textit{"The place was nice and calm"}. 
This situation is quite common in reality, increasing the difficulty of matching aspects with opinions. Nevertheless, current solutions either fail to capture these one-to-many relationships~\cite{DBLP:conf/emnlp/XuLLB20} or ignore the semantic relationship between word pairs in a triplet~\cite{wu-etal-2020-grid}. %Word pairs composed of elements within a triplet may have a semantic connection with each other, such as a dependency relationship. %fails to solve the one-to-many phenomenon. %A detailed comparision between~\citet{Peng_Xu_Bing_Huang_Lu_Si_2020} and~\citet{xu2020aspect} is shown in Table~\ref{tab_different_tag_schema}.  
%~\citet{wu-etal-2020-grid} address this issue with a Grid Tagging Schema (GTS) labeling the relationship of all word pairs, but they also ignores the semantic relationship between word pairs and applies an unstable inference strategy.

%One crucial observation is that there are various relationships between triples (syntactic dependence, semantically similar words connections), which have not been fully explored, resulting in the implicit relationship between words not being effectively used. As shown in Figure~\ref{aste_example}, there exists a dependency called $nsubj$ (nominal subject) between $waiters$ and $friendly$, which indicates that there may be an aspect. The two opinions $friendly$ and $so \ so$ in the sentence are also associated with each other, where there is a dependency called $conj$ (conjunct) implying they have similar attributes. Opinions usually modify aspects and analogous words tend to be related.
Furthermore, various relationships exist among triplets, such as syntactic dependence and semantic word similarity, which have been neglected. For example, as shown in Figure~\ref{aste_example}, there is a nominal subject dependency (called $nsubj$) between $waiters$ and $friendly$, indicating that there exists an aspect. Also, the two opinions, $friendly$ and $so \ so$ in the sentence are associated with each other, where there is a conjunct dependency (called $conj$), implying they have similar attributes. %Opinions usually modify aspects and analogous words tend to be related.

To fully utilize these implicit relationships, we design a \textit{\textbf{S}emantic and \textbf{S}yntactic \textbf{E}nhanced Aspect \textbf{S}entiment Triplet \textbf{E}xtraction model} (\modelname).  \modelname~utilizes semantic and syntactic information from words, which helps to distinguish words' attributes and identify the relationship between word pairs. In order to better leverage these relationships, we build a \textit{Graph Neural Network} (GNN) based model to capture the interactions between words and triplet elements.
For each sentence, we transform it into a unique text graph representation, where each node is a word, and the edges are established based on attention to the words themselves, adjacent relationships, and syntactic dependencies. Such a concise and effective text graph can obtain the precise meaning of each word and gain insight into their relations.

Moreover, we further utilize LSTM~\cite{doi:10.1162/neco.1997.9.8.1735} to learn the contextual semantics of each word from a sequential perspective, forming a Graph-Sequence duel modeling of a sentence. In this way, \modelname{} has an excellent ability to distinguish the categories of words and more accurately recognize the relationship between word pairs. 
With the semantic and syntactic enhanced module, the correlation between word pairs is well captured, yielding a more simple inference strategy for triplet extraction. Since \modelname{} can perceive the semantics and syntax from words excellently, we only need to infer once for all datasets to obtain more accurate triplets and save time overhead. Finally, we parse out the triplets from the final predictions. 
%Through our semantic and syntactic enhanced model, we output relatively accurate initial probability distributions between words. According to the initial probability distributions, if a word is predicted as an aspect and another word not far away is judged as an opinion, then they are likely to form a triplet. To take advantage of these potential connections from current prediction results, we only infer once from them for all dataset to obtain more accurate triplets, since our model can perceive the semantics and syntax from words excellently. This shows the efficiency of our inference strategy, which only requires a small time overhead. Finally, we parse out the triplets from the final predictions. 
%Since Graph Neural Network (GNN) can build relational structure and obtain global information~\citep{Yao_Mao_Luo_2019}, we use GNN to model the dependency and capture the connections between word pairs in advance, and also combine LSTM to learn contextual semantics. In this way, our model is able to utilize multifaceted features flexibly. 

We run extensive experiments on four benchmark datasets. The experimental results show that \modelname{} achieves significantly better performance than existing state-of-the-art approaches by fully exploiting the syntactic and semantic 
18
 relationships between word pairs.

To summarize, our main contributions include the following:
%\begin{itemize}
\squishlist
    \item We design a graph representation of a sentence which integrates the syntactic dependency, semantic relatedness, and positional relationship between words, and encode it with Graph Neural Networks to fully exploit the various correlations. %for aspect triplet sentiment extraction.
    \item We further model the sentence with LSTM to incorporate its sequential information, forming a Graph-Sequence duel modeling paradigm. Moreover, 
    %\item %By observing the characteristics of the previous prediction results, 
    we only need to infer once for all datasets, demonstrating the superiority of \modelname. %demonstrating the convenience of our proposed inference strategy and the effectiveness of our model.
    \item We make extensive experiments, and the results show \modelname~ outperforms all state-of-the-art approaches significantly for triplet extraction.
%\end{itemize}
\squishend

\section{Our Approach}
\begin{figure}[t]
\centering
\centerline{\includegraphics[width=\linewidth]{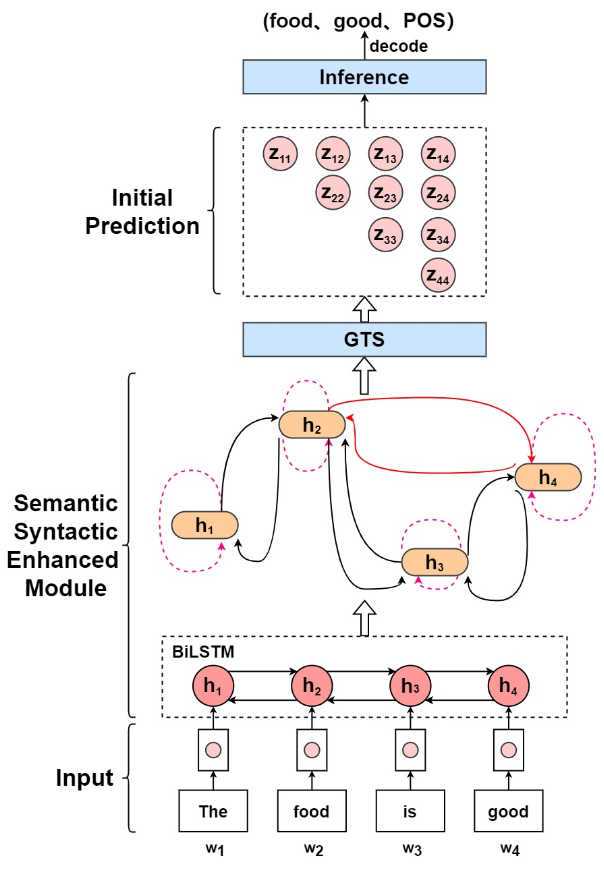}}
\caption{The overall architecture of our end-to-end model~\modelname{}. In our text graph, the type of dashed edges is self-loop, the type of black solid edges is neighbor edge, and the type of red solid edge is dependent edge.} 
\label{fig_model}
\end{figure}

We design an effective framework to complete triplet extraction in an end-to-end fashion. The overall model architecture is shown in Figure~\ref{fig_model}. In this section, we first define the ASTE task, describe the Grid Tagging Schema and deconstruct triplets from it in detail. We next present \modelname{} model, %\textbf{S}emantic and \textbf{S}yntactic \textbf{E}nhanced Aspect  \textbf{S}entiment Triplet \textbf{E}xtraction model (\modelname), 
followed by our inference strategy.

\subsection{Task Definition and Preliminaries}
\paragraph{Definition: Triplet Extraction.}
Given an input sentence $x=\left\{x_{1}, x_{2}, \cdots, x_{n}\right\}$ with length n, each word has two tag labels: the aspect tag label and the opinion tag label, respectively. Their tagging schema is $\mathcal{Y}$ = \{B, I, O\}, denoting the beginning, inside, outside of one aspect term or opinion term. Meanwhile, each aspect target is annotated with a sentiment polarity label $\mathcal{S}$ = \{NEU, POS, NEG\}, denoting neutral, positive, and negative sentiment expressed towards itself. Our goal is to extract a set of triplets $\mathcal{T}=\left\{(a, o, s)_{m}\right\}_{m=1}^{|\mathcal{T}|}$ from the sentence $x$, where the notations $a$, $o$, and $s$ stand for an aspect, an opinion, and corresponding sentiment polarity, respectively. The notation $(a, o, s)_{m}$ is a triplet in $x$ and $|\mathcal{T}|$ 
represents the total number of triplets in this sentence.

\begin{figure}[t]
\centerline{\includegraphics[width=\linewidth]{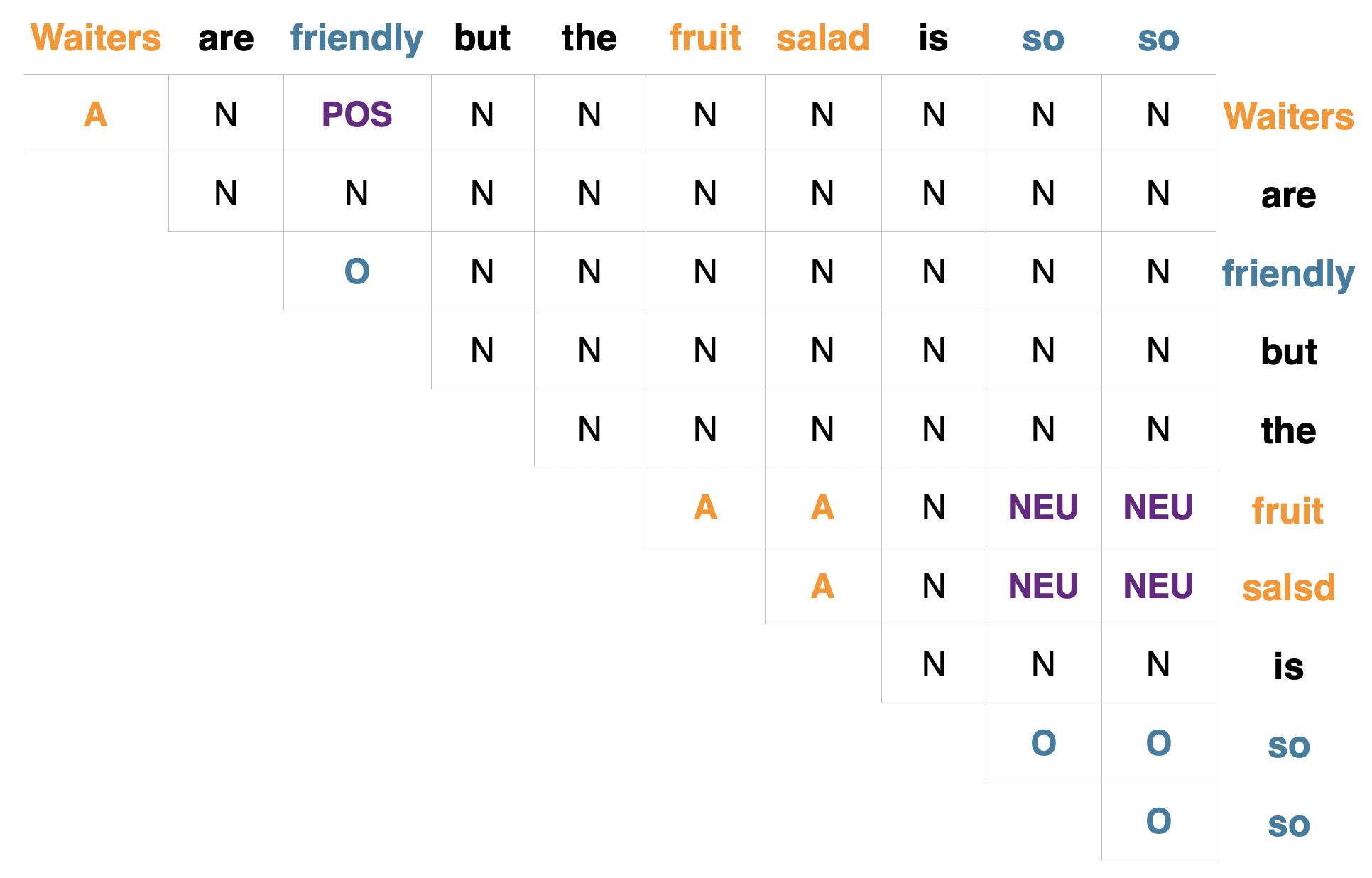}}
\caption{A tagging example for GTS } 
\label{fig1}
\end{figure}

\paragraph{Grid Tagging Schema.} To tackle the ASTE task, a \textit{Grid Tagging Schema} (GTS) was proposed by~\citet{wu-etal-2020-grid}, which adopts six tags $\mathcal{G}=\{\mathrm{A}, \mathrm{O}, \mathrm{NEG}, \mathrm{NEU}, \mathrm{POS}, \mathrm{N}\}$ to represent the relationship for any pair of two words $\left(w_{i}, w_{j}\right)$ in a sentence. The two tags, A and O, denote the word-pair $\left(w_{i}, w_{j}\right)$ is the same aspect or opinion, respectively. The three tags NEG, NEU, POS denote negative, neutral, or positive emotions expressed for the triplet consisting of the pair of words $\left(w_{i}, w_{j}\right)$ that exactly contains  an aspect term and an opinion term. The tag N denotes non above relations for word-pair $\left(w_{i}, w_{j}\right)$. A tagging example is shown in Figure~\ref{fig1}. In detail, the three coordinates in the grid $\left(5, 5\right)$, $\left(6, 6\right)$, and $\left(6, 5\right)$ respectively form word pairs $\left(fruit, fruit\right)$, $\left(salad, salad\right)$, and  $\left(fruit, salad\right)$, which are labeled A because they all belong to the same aspect. The same logic applies to opinions. The coordinate $\left(2, 0\right)$ is labeled POS because it makes a correct triplet $\left(Waiters, friendly, POS\right)$, which contains exactly the right aspect, opinion, and sentiment information. For simplicity, we use an upper triangular grid. 

\paragraph{Triplets Decoding.} we explain how to decode triplets based on the predicted grid tags. We take the decoding algorithm designed by ~\citet{wu-etal-2020-grid}. First, both aspects and opinions were identified using the predictive tags of all word pairs $\left(w_{i}, w_{j}\right)$ on the main diagonal without considering other word pairs' constraints. The span consisting of continuous A is regarded as a complete aspect, and the span consisting of continuous O is detected as a complete opinion. At this point, we have extracted the aspect $a$ and opinion $o$. Then, we count the predicted tags of all word pairs $\left(w_{i}, w_{j}\right)$ when $w_{i} \in a$ and $w_{j} \in o$. The most predictive sentiment label $s \in S$ is regarded as sentiment polarity for triplet $\left(a, o, s\right)$. When there are multiple most predictive sentiment labels, then the label is decided by the order: positive $\textgreater$ neutral $\textgreater$ negative. If they are all predicted to be label N, we consider that $a$ and $o$ cannot constitute a triplet.

\subsection{Semantic and Syntactic Enhanced ASTE Model}
Since this task requires extracting multiple elements from a sentence, it is important to design a model that can effectively distinguish the properties of words and master the relationship between them. \modelname~ first uses LSTM to encode sentences so that we can perceive contextual semantic. In order to capture many-sided features, \modelname~next applies graph neural network to model syntactic dependency, semantic relatedness, and positional relationship between words. Finally, an inference strategy is proposed, which only makes one inference to further extract more accurate triplets for all datasets.%, which in turn shows its advantages and the excellence of our model.
\subsubsection{Graph-Sequence Duel Representation}
We first apply a bidirectional \textit{Long Short Term Memory} (LSTM) networks ~\citep{doi:10.1162/neco.1997.9.8.1735} to encode the input sentence $x$. LSTM is capable of learning contextual semantic representation since it can mark key semantics from previous time steps. Hence, we learn contextual features $\left\{\mathbf{h}_{1}, \mathbf{h}_{2}, \cdots, \mathbf{h}_{n}\right\}$ for the input sequence. 

We observe that different words in a sentence often have various internal relationships. As elaborated in Figure~\ref{aste_example}, there is a syntactic dependency between $waiters$ and $friendly$, since opinions often modify aspects. Besides, words that are semantically similar may also be related. The two opinions, $friendly$ and $so \ so$, although they are far apart, there is still a dependency between them. Therefore, it is of great help to model the relationships and grasp semantic and syntactic information from words. With this in mind, we build a unique text graph for every input sentence using graph neural network.

Formally, a text graph $G=(V, E)$ is a structure used to represent words and their relations, which consists of the set of nodes $V$ and the set of edges $E$. Each word in the sentence is regarded as a node, while the relationships between words are considered edges. We construct three types of edges: self-loop edge, neighbor edge, and dependency edge. If there is an edge connecting to the node itself, then the edge is the self-loop edge. The edge connecting a node and its neighbor is a neighbor edge, while if there exists a dependency relationship between two nodes, then there is a dependency edge between them.
%Information carried by each node itself is vital, so each node is connected to its own. This type of edge is called self-loop. Next, it is very effective for each node to gather semantics from adjacent nodes, because they supplement the context. We connect each node to adjacent nodes and this type of edge is called neighbor edge. Finally, an edge exists between two nodes if they have a dependency relation. We call this kind of edge dependent edge, which enables each node obtain information from related nodes. 
Specifically, we define the text graph as follows:
\begin{equation}
    V=\left\{v_{i} \mid i \in[1, n]\right\}
\end{equation}
\begin{equation}
   E=\left\{e_{ij} \mid j=[i-1,i+1] \cup D_i\right\}
\end{equation}
where $D_i$ represents a set of nodes with which node $v_i$ has a dependency. All edges are bidirectional and the node feature for $v_i$ is taken from $\mathbf{h}_{i}$. We adopt GraphSAGE~\citep{hamilton2017inductive} to generate representations $\left\{\widetilde{\mathbf{h}}_{1}, \widetilde{\mathbf{h}}_{2}, \cdots, \widetilde{\mathbf{h}}_{n}\right\}$ for each node. We chose LSTM aggregator from GraphSAGE because it has stronger expressive ability. %Up to this step, the representations we obtain has gathered many aspects of semantics.

Then, we concatenate the integrated representations of $w_i$ and $w_j$ to represent all word pairs $\left(w_{i}, w_{j}\right)$, i.e., $\mathbf{r}_{i j}=\left[\widetilde{\mathbf{h}}_{i} ; \widetilde{\mathbf{h}}_{j}\right]$, where $[\cdot ;]$ is a concatenation operation.
All representations of word pairs correspond to cells in our grid, which is then fed to a linear layer to calculate initiatory probability distribution $\mathbf{z}_{i j} \in \mathbb{R}^{|\mathcal{G}|}$
through:
\begin{equation}
\mathbf{z}_{i j}=\mathbf{W}_{s} \mathbf{r}_{i j}+\mathbf{b}_{s}
\end{equation}
where $\mathbf{W}_{s}$ and $\mathbf{b}_{s}$ are trainable parameters.

\subsubsection{Inference Strategy}
%Different elements of ASTE mutually benefit each other. Therefore, ~\citet{wu-etal-2020-grid} designed an effective inference strategy to take advantage of these potential signs to promote ASTE. 
The initial probability distribution $\mathbf{z}_{ij}$ between all word pairs obtained above can further facilitate more accurate extraction of triplets. For instance, if $\left(0, 0\right)$ and $\left(2, 2\right)$ in grid tagging example are predicted to be A and O, respectively, then the position at which they intersect $\left(0, 2\right)$ is even less likely to be predicted to be N, and vice versa. Also, since many aspects or opinions are made up of multiple words, if a certain coordinate is predicted as one of $\mathcal{S}$, then its adjacent locations are more likely to be predicted to be the same sentiment label. 

Therefore, we employ an inference strategy to obtain more accurate triplets by observing the characteristics of the initial probability distributions through the below processes. Formally, new feature representation $\mathbf{g}_{i j}$ learning is as follows:
\begin{equation}
\begin{array}{c}
     \mathbf{z}_{i}=\operatorname{maxpooling}\left(\mathbf{z}_{i,:}\right)
     \\
     \\
     \mathbf{z}_{j}=\operatorname{maxpooling}\left(\mathbf{z}_{j,:}\right)
     \\
     \\
     \mathbf{\widetilde{r}}_{i j}=\left[\mathbf{r}_{i j} ; \mathbf{z}_{i} ; \mathbf{z}_{j} ; \mathbf{z}_{i j}\right]
     \\
     \\
     \mathbf{g}_{i j}=\mathbf{W}_{g} \mathbf{\widetilde{r}}_{i j}+\mathbf{b}_{g}
\end{array}
\end{equation}
\noindent where $\mathbf{W}_{g}$ and $\mathbf{b}_{g}$ are trainable parameters. The symbol $[\cdot ;]$ represents a concatenation operation. Concretely, $\mathbf{z}_{i,:}=\left(\mathbf{z}_{1: i, i}, \mathbf{z}_{i, i: n}\right)$ because of the upper triangular grid in GTS. $\mathbf{z}_{i}$/$\mathbf{z}_{j}$ works by capturing the associated features between $w_i$/$w_j$ and other words.  %and the efficiency of the inference strategy. 

It is worth noting that inference strategy by~\citet{wu-etal-2020-grid} are unable to well capture the relationship between words, thus yielding indefinite number of iterations for inference, which increases the time complexity when the number of inferences is large. %Besides, the number of inferences remains uncertain when the dataset changes.
In contrast, we only need to infer once for all datasets with semantic and syntactic enhanced module, which further proves the superiority of \modelname.

Finally, we send $\mathbf{g}_{i j}$ to a linear layer with softmax activation function for classification.
\begin{equation}
p_{i j}=\operatorname{softmax}\left(\mathbf{W}_{p} \mathbf{g}_{i j}+\mathbf{b}_{p}\right)
\end{equation}
where $\mathbf{W}_{p}$ and $\mathbf{b}_{p}$ are trainable parameters. 

\subsection{Training Loss Function}
The training goal for the ASTE task is to minimize the cross-entropy error for all word pairs. The unified loss function is defined as: 
\begin{equation}
   \mathcal{L}=-\sum_{i=1}^{n} \sum_{i=i}^{n} \sum_{k \in \mathcal{G}} \mathbb{I}\left(y_{i j}=k\right) \log \left(p_{i, j}\right)
\end{equation}
where $y_{i j}$ denotes the one-hot vector of ground truth for the word pair $\left(w_{i}, w_{j}\right)$ and $\mathbb{I}(\cdot)$  indicates the k-th component being 1.

\begin{table*}
\centering
\caption{ Statistics of datasets (\#S, \#T, \#-, \#0, and \#+ denote number of sentences, triplets, negative triplets, neutral triplets, and positive triplets respectively.)}
\resizebox{15cm}{1.0cm}{
\begin{tabular}{c|ccccc|ccccc|ccccc|ccccc}
   \hline 
   \multirow{2}{*}     { Dataset } & 
   \multicolumn{5}{c|} { 14res } & 
   \multicolumn{5}{c|} { 14lap } & 
   \multicolumn{5}{c|} { 15res } & 
   \multicolumn{5}{c}  { 16res } \\
    \cline { 2 - 21 } 
    & \#S & \#T & \#- & \#0 & \#+
    & \#S & \#T & \#- & \#0 & \#+
    & \#S & \#T & \#- & \#0 & \#+
    & \#S & \#T & \#- & \#0 & \#+ \\
\hline
\textbf{train} & 1259 & 2356 & 491 & 172 & 1693 
& 899 & 1452 & 533 & 111 & 808 
& 603 & 1038 & 210 & 29 & 799
& 863 & 1421 & 330 & 55 & 1036 \\
\textbf{val} & 315 & 580 & 107 & 46 & 427 
& 225 & 383 & 136 & 48 & 199 
& 151 & 239 & 49 & 9 & 181 
& 216 & 348 & 77 & 8 & 263 \\
\textbf{test} & 493 & 1008 & 156 & 68 & 784 
& 332 & 547 & 116 & 67 & 364 
& 325 & 493 & 144 & 25 & 324 
& 328 & 525 & 79 & 30 & 416 \\
\hline
\end{tabular}}
\label{tab:dataset_stat}
\end{table*}

\section{Experiments}
\subsection{DataSets}
We conduct experiments on four datasets integrated by \citet{wu-etal-2020-grid}.
Each dataset has been divided into three parts: training set, validation set, and test set. 
Table \ref{tab:dataset_stat} lists the statistics for these datasets. 14res, 15res, and 16res belong to the restaurant domain, while 14lap is of laptop domain. Each sentence has been annotated with a sequence of aspect tags and opinion tags and sentiment polarity of corresponding aspects. These datasets originally come from SemEval Challenges \cite{pontiki-etal-2014-semeval,pontiki-etal-2015-semeval,pontiki-etal-2016-semeval}.

Note that each sentence may have more than one aspect and opinion. Besides, one aspect may be associated with multiple opinions and vice versa. For 14res, 14lap, 15res, and 16res, the proportion of one-to-many data reaches 37.27\%, 38.54\%, 33.39\%, and 33.13\%, respectively.
Various relationships usually exist between aspects and opinions, using them is beneficial to triplet extraction. We count the ratio of triplets with implicit relationships. For these four datasets, they are 79.37\%, 74.22\%, 76.27\%, and 80.57\%, respectively. 

\subsection{Baselines}
We compare the performance of \modelname{} with the following approaches, where most triplet extraction models currently are done in a pipeline manner, and few state-of-the-art models are in an end-to-end way.
%\begin{itemize}
\squishlist
\item \textbf{Peng-unified-R+PD.} \citet{Peng_Xu_Bing_Huang_Lu_Si_2020} proposed a pipeline approach in two stages. The first stage model (Peng-unified-R) jointly extracts aspects with sentiment using the unified tagging schema and opinion location in the BIEOS tagging schema. It leverages mutual information between aspects and opinions. In the second stage, all candidate triplets are generated, and a MLP-based classifier (PD) is applied to determine whether each triplet is valid or not.
\item \textbf{Li-unified-R+PD.} A pipeline approach combined by \citet{Peng_Xu_Bing_Huang_Lu_Si_2020}. In the first stage, the model~\citep{li2019unified} is modified to co-extract aspects with sentiment as well as extracting opinion. In the second stage, it applies the same classifier (PD) mentioned above to obtain all the valid triplets.
\item \textbf{Peng-unified-R+IOG.} A pipeline approach combined by ~\citet{wu-etal-2020-grid}. It first employs the model Peng-unified-R of \citet{Peng_Xu_Bing_Huang_Lu_Si_2020} for extracting aspects with sentiment, then uses IOG \citep{fan2019target} to produce final triplets. The IOG encodes the information from a given asepct to extract its opinion words.
\item \textbf{IMN+IOG.} Another pipeline approach combined by ~\citet{wu-etal-2020-grid}. It first employs the IMN \citep{he-etal-2019-interactive} for extracting aspects with sentiment, then uses the IOG \citep{fan2019target} to produce final triplets.
\item \textbf{Grid.} A state-of-the-art approach model proposed by ~\citet{wu-etal-2020-grid}, which designs a grid tagging schema to address triplet extraction in an end-to-end way. It employs an inference strategy to utilize the mutual indications between different opinion factors. For a fair comparison, we choose their model Grid-CNN and Grid-BiLSTM, which use CNN encoder and BiLSTM encoder respectively. 
%\end{itemize}
\squishend

\begin{table*}
\centering
\caption{ Experimental results of triplet extraction. Best results are in bold. The mark "*" means that \modelname{} significantly outperforms all baselines. The mark "-" means that the original code of the IMN method does not contain the resources required to run on the dataset 16res.}
\resizebox{15.5cm}{2.0cm}{
\begin{tabular}{c|ccc|ccc|ccc|ccc}
   \hline 
   \multirow{2}{*}     { Model } & 
   \multicolumn{3}{c|} { 14res } & 
   \multicolumn{3}{c|} { 14lap } & 
   \multicolumn{3}{c|} { 15res } & 
   \multicolumn{3}{c}  { 16res } \\
    \cline { 2 - 13 } 
    & P & R & F
    & P & R & F
    & P & R & F
    & P & R & F \\
\hline
\text { Li-unified-R+PD } 
& 41.44 & 68.79 & 51.68 
& 42.25 & 42.78 & 42.47 
& 43.34 & 50.73 & 46.69 
& 38.19 & 53.47 & 44.51 \\
\text { Peng-unified-R+PD }
& 44.18 & 62.99 & 51.89 
& 40.40 & 47.24 & 43.50 
& 40.97 & 54.68 & 46.79 
& 46.76 & 62.97 & 53.62 \\
\text {Peng-unified-R+IOG} 
& 58.89 & 60.41 & 59.64 
& 48.62 & 45.52 & 47.02 
& 51.70 & 46.04 & 48.71 
& 59.25 & 58.09 & 58.67 \\
\text {IMN+IOG} 
& 59.57 & 63.88 & 61.65 
& 49.21 & 46.23 & 47.68 
& 55.24 & 52.33 & 53.75 
& $-$ & $-$ & $-$ \\
Grid-CNN 
& 70.79 & 61.71 & 65.94 
& 55.93 &47.52 & 51.38 
& 60.09 & 53.57 & 56.64 
& 62.63 & 66.98 & 64.73 \\
Grid-BiLSTM 
& 67.28 & 61.91 & 64.49 
& 59.42 & 45.13 & 51.30 
& 63.26 & 50.71 & 56.29 
& 66.07 & 65.05 & 65.56 \\
\hline
\modelname{}
& 69.08 & 64.55 & \textbf{66.74}$^{*}$
& 59.43 & 46.23 & \textbf{52.01}
& 61.06 & 56.44 & \textbf{58.66}$^{*}$
& 71.08 & 63.13 & \textbf{66.87}$^{*}$ \\
\hline
\end{tabular}}
\label{tab:experiment_result}
\end{table*}

\subsection{Implementation Details}
Following the previous work~\citep{wu-etal-2020-grid}, we combine a 300-dimension domain-general embedding from GloVe~\citep{pennington-etal-2014-glove} and pre-trained with 840 billion tokens and a 100-
dimension domain-specific embedding trained with fastText~\citep{TACL999} to initialize double word embeddings for \modelname. The learning rate is 0.001, and the dropout rate is 0.5. We use Adam~\citep{DBLP:journals/corr/KingmaB14} as \modelname{} optimizer. The number of layer for LSTM is 1 and the cell is set to 50. The aggregator type from GraphSAGE we chose is LSTM. We use Stanza~\citep{qi2020stanza} to parse the dependencies in the sentence. The batch size is set to 32 for all datasets and the valid set is used for early stopping. We select the best model according to the best F1 score on the valid set and run the test set with it for evaluation.

Following previous work, we report experimental results based on precision (P), recall (R), and F1 scores. Note that the F1 score measures the performance of mating triplets, which means a triplet is correct only when the aspect span, its corresponding sentiment, and opinion span are all proper.

\subsection{Main Results For Triplet Extraction}
Table~\ref{tab:experiment_result} presents the main results of the final triplet extraction. \modelname{} surpasses all baselines significantly on all datasets. Compared with the best results of existing baselines, \modelname{} still achieves an apparent absolute F1 scores increase of 2.02\% and 1.31\% on 15res and 16res, respectively, and achieved an impressive increase of 0.80\% and 0.63\% on 14res and 14lap, respectively. Except for Grid-CNN and Grid-BiLSTM, the other models are all pipeline methods.

The experimental results show that~\modelname{} is far beyond these methods, which also strongly proves the advantages of the semantic and syntactic enhanced model. When we compare \modelname{} with competitive baselines, Grid-CNN and Grid-BiLSTM in detail, we find that the reason why we perform better on 14res and 15res is because we extract a more complete set of triplets in these two datasets, resulting a more significant recall.  The reason why we perform better on 14lap and 16res is because we extract more accurate triplets, resulting a more significant precision. Such comprehensive results demonstrate the strength of~\modelname{}, which has the ability to learn multi-faceted semantics and and is good at extracting triplets.

\section{Experiment Analysis}
\subsection{Ablation Study}
To investigate the effectiveness of different modules in \modelname, we conduct ablation study for the ASTE task. As shown in Table~\ref{tab:ablation_study}, \modelname{} represents our full model that equipped with all modules. Next, we will carefully observe the role of each module by introducing four model variants, namely Dep, Infer, Graph, and BiLSTM.

Infer means removing the inference strategy from \modelname. We can see that F1 scores drop sharply, which shows that the inference strategy can grasp the relationship between the three elements in the triplets from the previous round of predictions to promote the ASTE task. 
Dep means that when constructing a text graph for a sentence, we do not add the third edge type mentioned above. We can see that F1 scores drop except for res14, showing that overall the dependent edges can help the model better master relationships. The training set of 14res is larger than other datasets. When training the full model, we may overfit due to the setting of parameters (e.g., epoch, batch size), resulting in slightly lower performance, compared with Dep.

Graph means removing the graph-based GNN modules. After removing the entire graph, the performance of the model is greatly reduced. Obviously, the graph neural networks can well perceive the relational semantics and distinguish the characteristics of the words. The F1 scores also decline sharply when we remove the BiLSTM, which shows that contextual semantic information is helpful. Comparing Graph and BiLSTM, we find that the former has higher results on 14lap and 16res. It may be that these two datasets are more dependent on contextual semantic features. 
In general, each module of \modelname{} contributes to the extraction of triplets.

\begin{table}
\centering
\caption{ Results of ablation study for the ASTE task }
\begin{tabular}{c|c|c|c|c}
\hline \multirow{2}{*} { Models } & 14res & 14lap & 15res & 16res \\
\cline { 2 - 5 } & F & F & F & F \\
\hline 
\modelname & 66.23 & 52.01 & 58.66 & 66.87 \\
\hline
Infer & 64.20 & 48.68 & 56.90 & 63.27 \\
Dep & 66.74 & 50.43 & 57.43 & 64.98 \\
Graph & 62.12 & 46.37 & 53.77 & 63.63 \\
BiLSTM & 62.48 & 44.78 & 54.38 & 61.54 \\
\hline
\end{tabular}
\label{tab:ablation_study}
\end{table}

\begin{table}
\centering
\caption{ Results of triplet extraction on different aggregators and number of graph network layers }
\resizebox{7.5cm}{1.5cm}{
\begin{tabular}{c|c|c|c|c|c}
\hline \multirow{2}{*} { Aggregator} & \multirow{2}{*} { Layers } & 14res & 14lap & 15res & 16res \\
\cline { 3 - 6 } & & F & F & F & F\\
\hline \multirow{2}{*} { LSTM } 
 & 2 & 64.83 & 47.32 & 55.84 & 62.96 \\
 & 3 & 66.23 & 52.01 & 58.66 & 66.87 \\
\hline \multirow{2}{*} { Mean } 
 & 2 & 64.28 & 47.00 & 55.15 & 62.73 \\
 & 3 & 64.43 & 50.26 & 54.10 & 63.70 \\
\hline
\end{tabular}}
\label{tab:effect_aggre_type_and_layer_num}
\end{table}

\subsection{Effects of Aggregator Types}
In order to study the impact of aggregator types on performance, we report the results of different aggregator types for the ASTE task on these four datasets in Table~\ref{tab:effect_aggre_type_and_layer_num}. There are two types of aggregators, LSTM and Mean, adopted from~\citep{hamilton2017inductive}. The former is based on the LSTM structure~\citep{doi:10.1162/neco.1997.9.8.1735} and is applied to the random arrangement of the node's neighbors. The latter is just based on the mean operation. As shown in Table~\ref{tab:effect_aggre_type_and_layer_num}, when the network layers of the two aggregators are equal, no matter how many layers, the effect of the LSTM aggregator is better than that of the Mean aggregator. This phenomenon indicates that the LSTM aggregator has stronger expressive ability and is more suitable for the ASTE task.

\subsection{Effects of Graph Network Layers}
To examine the effects of the number of graph network layer, we also present the results of different layers on these four datasets to extract triplets. It can be observed that the experimental performance increases as the number of layers increases from 2 to 3 for the same type of aggregator. This proves that the ability of graph neural networks to gather features is related to the number of network layers. We notice that when the number of layers is set to 2, the LSTM aggregator has higher performance than the Mean aggregator by 0.55\%, 0.32\%, 0.69\%, and 0.23\% on the four datasets, respectively. Nevertheless, when the number of layers is 3, their performance differs by 1.80\%, 1.75\%, 4.56\%, and 3.17\%. As the number of layers increases, the performance gap between the LSTM aggregator and the Mean aggregator widens significantly, which further illustrates the advantage of the LSTM aggregator.

\begin{table*}[t]
\caption{ Case analysis. The first column is five representative examples, the second column is golden truth, and the other columns are the output results of different models.}
\resizebox{15.5cm}{2.1cm}{
\scriptsize
\renewcommand\arraystretch{1.2}
\setlength{\tabcolsep}{0.3mm}
%\resizebox{\textwidth}{15mm}
\centering
%\small
{
\begin{tabular}{c|c|c|c|c}
\hline Example & Golden Triplets & Grid-BiLSTM & Grid-CNN & Our\\
\hline
\hline
The bread is top notch as well 
& (bread,top notch,POS) 
& (bread,top notch,POS)
& (bread,top notch,POS)
& (bread,top notch,POS)\\
\hline
The staff should be & \multirow{2}{*}{(staff,friendly,NEG)} & (staff,more friendly,POS) & \multirow{2}{*} {(staff,more friendly,POS) }& \multirow{2}{*} {(staff,friendly,NEG)}\\
a bit more friendly & & (staff,should be,POS) & &\\
\hline
Made interneting difficult to maintain & (interneting,difficult,NEG) & (maintain,difficult,POS) & 
(maintain,difficult,POS) & 
(interneting,difficult,NEG)\\
\hline
It has so much more speed 
& (speed,much more,POS) & 
\multirow{2}{*}{(screen,sharp,POS)} &  \multirow{2}{*}{(screen,sharp,POS)} &
(screen,sharp,POS) \\
and the screen is very sharp & (screen,sharp,POS) & & & (speed,more,POS)\\
\hline
The food was & 
(food,tasty,POS) & 
(food,tasty,POS) & 
(food,tasty,POS) &
(food,tasty,POS)\\
extremely tasty , & 
(food,creatively presented,POS) & (food,creatively presented,POS) &
(food,creatively presented,POS) &
(food,creatively presented,POS)\\
creatively presented and & (wine,excellent,POS) & 
(wine,excellent,POS) &
(wine,excellent,POS) &
(wine,excellent,POS)\\
the wine excellent & & (food,excellent,POS) & &\\
\hline
\end{tabular}}}
\vspace{-3mm}
\label{tab_case_analysis}
\end{table*}

\subsection{Case Study}
Five typical cases are presented in Table~\ref{tab_case_analysis}. The first example is a simple case without complicated word order and all models can predict accurately. The second example comes from the restaurant field, which expresses a negative attitude tactfully. Both Grid-BiLSTM and Grid-CNN incorrectly predict sentiment for "staff", and Grid-CNN mistakenly predicts "should be" as an aspect.

The third example directly expresses negative sentiment, which is picked from the laptop field. We can observe that Grid-LSTM and Grid-CNN mistakenly regard "maintain" as an aspect, and also make a false prediction for sentiment. For these two examples, \modelname{} makes accurate judgments, which shows that \modelname{} can better understand the context and distinguish the characteristics of words. 

There are 2 triplets in the fourth example. All methods extract the triplet containing "screen". Unlike other models, \modelname{} successfully identifies the second aspect "speed" and its sentiment. Though lacking of an opinion word "much", \modelname{} has stronger recognition ability.

The last one is a more complicated example with 3 triplets, where an aspect corresponds to multiple opinions. We see that Grid-BiLSTM mistakenly matches "food" and "excellent" as a triplet. Both Grid-CNN and \modelname{} make correct predictions. In general, the above analysis further proves that \modelname{} can better understand the semantics and recognize the relationship more accurately.

\section{Related Work}
ASTE originates from another highly concerned research topic called \textit{Aspect Based Sentiment Analysis} (ABSA) ~\cite{pontiki-etal-2014-semeval,pontiki-etal-2015-semeval,pontiki-etal-2016-semeval}. 
The research process of ABSA can be divided into three stages. 

\paragraph{Separate Extraction.} Traditional studies have divided ABSA into three subtasks, namely, \textit{aspect extraction} (AE), \textit{opinion extraction} (OE), and \textit{aspect sentiment classification} (ASC). The AE task \cite{DBLP:conf/ijcai/YinWDXZZ16,DBLP:conf/ijcai/LiBLLY18,xu2018double,ma2019exploring} requires the extraction of aspects, while the OE task’s goal \citep{fan2019target} is to identify opinions expressed on them. The ASC task has attracted much more attention, which refers to classifying sentiment polarity for a given aspect target \cite{yang2017attention,chen2017recurrent,AAAI1816541,li2018transformation,xue2018aspect,wang-etal-2018-target,li2019exploiting} because the sentiment element carries crucial semantic information for a text.
\citet{zhang2019aspect} develops \textit{aspect-specific Graph Convolutional Networks} (ASGCN) that integrates with LSTM for the ASC task. Compared with ASGCN, \modelname{} has richer edge types and fewer training parameters. Since its aspect-specific structure must depend on the given aspect, ASGCN lacks scalability and cannot be extended to triplet extraction in an end-to-end fashion.
Besides, solving these three subtasks individually lacks practical application value and ignores the internal relation between them.

\paragraph{Pair Extraction.} Recently, many studies have proposed effective models to jointly extract aspects and their sentiments \cite{zhang2015neural,AAAI1714931,li2019learning,li2019exploiting,li2019unified}. %However, these methods use the unified tagging schema, leading to sentiment inconsistency.
\citet{hu2019open} design a Span-Based method but conclude the pipeline model is better than the unified model. There is also a practice to co-extract aspects and opinions \cite{wang2017coupled,dai2019neural}. These pair extraction models still cannot fully understand a complete picture regarding sentiment and dig deeper into the interconnections between subtasks.

\paragraph{Triplet Extraction.}  The ASTE task is more challenging and application value. \citet{Peng_Xu_Bing_Huang_Lu_Si_2020} first propose a two-stage model for ASTE, which in the first stage co-extracts aspects with the associated sentiment and finishes opinion extraction in the form of a standard sequence labeling task. The second stage employs a binary classifier to match aspects and opinions to obtain final triplets. Following this work, \citet{DBLP:conf/emnlp/XuLLB20} employ a model with a position-aware tagging scheme to extract a triplet jointly, but it cannot apply to the one-to-many phenomenon. \citet{wu-etal-2020-grid} design a novel grid tagging schema to address triplet extraction, but their end-to-end model ignores the dependencies among words. Besides, the inference rounds of their inference strategy are not unified for each dataset, which may cause instability and high time complexity if the rounds rise. 

\section{Conclusion}
\textit{Aspect Sentiment Triplet Extraction} (ASTE) requires extracting aspects, corresponding opinions, and sentiment from user reviews. Different from previous work, we take advantage of multiple semantic relationships between word pairs and effectively capture the inner connection between such three elements. In this paper, we construct a novel model with a relational structure by creating a unique text graph for each sentence using \textit{Graph Neural Network} (GNN). We also combine LSTM to obtain contextual semantics. Through the above mentioned rich structure, \modelname{} can understand the context well and effectively recognize the identify between words. Besides, the inference strategy becomes more efficient because it only needs to be inferred once for all datasets, reducing the time complexity. Our end-to-end model achieves state-of-the-art performance on all datasets for triplet extraction. Experimental results show that \modelname{} remarkably captures the connection between word pairs and recognizes their relationship.

\section*{Acknowledgments}
This work is supported by the National Key Research and Development Program of China under Grant (No. 2020AAA0108501).

%The acknowledgments should go immediately before the references. Do not number the acknowledgments section.
%\textbf{Do not include this section when submitting your paper for review.}

\bibliographystyle{acl_natbib}
\bibliography{acl2021}

%\appendix

\end{document}